# Evaluation of k-means time series clustering based on z-normalization and NP-Free

Ming-Chang Lee[1], Jia-Chun Lin[2] and Volker Stolz[3]

[1,3]Department of Computer science, Electrical engineering and Mathematical sciences, Høgskulen på Vestlandet (HVL), Bergen, Norway

[2]Department of Information Security and Communication Technology, Norwegian University of Science and Technology,
Gjøvik, Norway

[1] mingchang1109@gmail.com
[2] jia-chun.lin@ntnu.no
[3] volker.stolz@hvl.no

28th January 2024

# Evaluation of k-means time series clustering based on z-normalization and NP-Free

Ming-Chang Lee[1][a], Jia-Chun Lin[2][b], and Volker Stolz[3][c]

[1,3]*Department of Computer science, Electrical engineering and Mathematical sciences, Høgskulen på Vestlandet (HVL), Bergen, Norway*
[2]*Department of Information Security and Communication Technology, Norwegian University of Science and Technology (NTNU), Gjøvik, Norway*
*mingchang1109@gmail.com[1], jia-chun.lin@ntnu.no[2], volker.stolz@hvl.no[3]*

Abstract: Despite the widespread use of k-means time series clustering in various domains, there exists a gap in the literature regarding its comprehensive evaluation with different time series normalization approaches. This paper seeks to fill this gap by conducting a thorough performance evaluation of k-means time series clustering on real-world open-source time series datasets. The evaluation focuses on two distinct normalization techniques: z-normalization and NP-Free. The former is one of the most commonly used normalization approach for time series. The latter is a real-time time series representation approach, which can serve as a time series normalization approach. The primary objective of this paper is to assess the impact of these two normalization techniques on k-means time series clustering in terms of its clustering quality. The experiments employ the silhouette score, a well-established metric for evaluating the quality of clusters in a dataset. By systematically investigating the performance of k-means time series clustering with these two normalization techniques, this paper addresses the current gap in k-means time series clustering evaluation and contributes valuable insights to the development of time series clustering.

## 1 INTRODUCTION

Time series clustering is a data mining technique that involves grouping similar time series into clusters without prior knowledge of these cluster definitions. To elaborate, clusters are established by aggregating time series with significant similarity to other time series within the same cluster while maintaining minimal similarity with time series in different clusters (Aghabozorgi et al., 2015).

In recent years, there has been an increasing need for time series clustering due to the explosion of the Internet of Things (IoT) in diverse areas. Vast amounts of time series data are continuously measured and collected from connected devices and sensors, and they often require clustering and analysis. Various clustering approaches have been introduced and employed to address this demand, including k-means (MacQueen et al., 1967), hierarchical clustering (Kaufman and Rousseeuw, 2009), k-Shape (Paparrizos and Gravano, 2015), Kernel K-means (Dhillon et al., 2004), etc. Among these, k-means is one of the most popular and widely used techniques, known for its simplicity and efficiency in partitioning time series data into distinct clusters (Ruiz et al., 2020). However, clustering raw time series can be challenging because the scales and magnitudes of different time series may vary significantly. Hence, it becomes necessary to apply normalization techniques before clustering the raw time series data.

z-normalization is one of the most commonly used normalization approach for time series (Dau et al., 2019), and it is widely employed by many representation approaches and clustering approaches because of its simplicity and effectiveness. However, z-normalization may cause certain distinct time series to become indistinguishable (Lee et al., 2023a), therefore misleading clustering approaches or representation approaches, which in turn has a negative impact on the performance of clustering results and the representation approaches.

NP-Free is a real-time time series representation approach introduced by Lee et al. (Lee et al., 2023b).

---

[a]  https://orcid.org/0000-0003-2484-4366
[b]  https://orcid.org/0000-0003-3374-8536
[c]  https://orcid.org/0000-0002-1031-6936

NP-Free possesses the unique capability to dynamically transform any raw time series into a root-mean-square error (RMSE) series in real time. This transformation ensures that the resulting RMSE series serves as an effective representation of the original raw series. Notably, NP-Free's elimination of preprocessing steps, such as z-normalization, distinguishes it from conventional representation approaches. This characteristic enables NP-Free to be an alternative normalization approach to z-normalization in time series clustering applications.

Considering these two distinct options for time series normalization, the impact of z-normalization and NP-Free on the performance of k-means time series clustering, especially concerning its clustering quality, remains unknown in the literature. Therefore, in this paper, we aims to fill this gap. Our goal is to analyze and compare the performance of k-means time series clustering when applied with z-normalization and NP-Free, providing insights into how each normalization technique influences the quality of clusters generated by k-means time series clustering.

Two experiments, utilizing real-world open-source time series datasets from the UEA&UCR archive (Dau et al., 2019), were conducted. The clusters generated by the two variants of k-means time series clustering were assessed using the silhouettes score, which is a metric used to evaluate the quality of clusters in a dataset (Rousseeuw, 1987). The experiment results shows that the cluster quality is significantly influenced by z-normalization and NP-Free. Our evaluation analysis valuable insights to the development of time series clustering.

The rest of the paper is organized as follows: Section 2 presents background information about k-means time series clustering, z-normalization, and NP-Free. In Section 3, we introduce related work. Section 4 provides evaluation details and discusses the experiments along with the corresponding results. Finally, Section 5 concludes this paper and outlines future work.

## 2 BACKGROUND

This section introduces k-means time series clustering, z-normalization, and NP-Free.

### 2.1 k-means time series clustering

k-means time series clustering is a unsupervised machine learning approach designed to group time series into distinct clusters. The method, its initial formulation, was first introduced by Mac Queenin1967 (Mac-Queen et al., 1967), and the approximation developed by Lloyd in 1982 (Lloyd, 1982) has proven to be more popular in application. It is widely used due to its ease of implementation, simplicity, and efficiency.

k-means minimizes the distance between each time series and the centroid of its assigned cluster, with distances computed using various metrics such as Euclidean distance or dynamic time warping (DTW) distance.

Before initiating k-means clustering, two parameters must be determined: the number of clusters, denoted by $k$, and the initial centroids. While a fixed parameter configuration yields a consistent clustering result, it is important to note that different configurations typically lead to varying outcomes. Consequently, a common approach is to execute k-means multiple times with different parameter configurations and subsequently select the best clustering outcome.

### 2.2 z-normalization

z-normalization (also known as z-score normalization) is a statistical technique used in data processing and analysis. It transforms data into a standard scale with a mean of 0 and a standard deviation of 1 (Senin, 2016). The purpose of z-normalization is to simplify the interpretation and comparison of different datasets, making them directly comparable. z-normalization is typically applied to individual data points within a dataset by subtracting the mean of the dataset from each data point and then dividing the result by the standard deviation. The formula of z-normalization is shown below.

$$\bar{z}_i = \frac{z_i - \mu}{\sigma} \quad (1)$$

where $z_i$ is the $i$-th data point in a time series, $\mu$ is the mean of all data points in the time series, $\sigma$ is the standard deviation of all the data points, and $\bar{z}_i$ is the z-normalized value (i.e., z-score) of the the $i$-th data point derived from the formula.

z-normalization is considered an essential preprocessing step for time series representation approaches because it allows these approaches to focus on the structural similarities/dissimilarities of time series rather than on the original data point values (Senin, 2016). However, z-normalization has some limitations and drawbacks, including sensitivity to outliers, loss of the original data scale, lack of invariance. Additionally, when applied to flat time series, it can amplify fluctuations, such as noises, resulting in an negative impact on data mining techniques, such as Matrix Profile (Paepe et al., 2019).

## 2.3 NP-Free

NP-Free is a real-time time series representation approach introduced by Lee et al. (Lee et al., 2023a). It eliminates the need for preprocessing input time series using z-normalization and does not require users to tune any parameter in advance. NP-Free directly generates a representation for a raw time series by transforming the time series into a root-mean-square error (RMSE) series in real time. Due to the no preprocessing need, NP-Free can serve as an alternative normalization approach to z-normalization in clustering applications.

NP-Free adopts Long Short-Term Memory (LSTM) and the Look-Back and Predict-Forward strategy used by RePAD (Lee et al., 2020) to generate representations for time series. More specifically, NP-Free keeps predicting the next data point in the target time series based on three historical data points and keeps calculating the corresponding RMSE value between the observed and predicted data points. By doing so, NP-Free can convert the target time series into a RMSE series.

Figure 1 illustrates the algorithm of NP-Free where $t$ denotes the current time point starting from 0 when NP-Free is applied to a raw time series. Let $c_t$ be the real data point collected at time point $t$, and $\widehat{c_t}$ be the data point predicted by NP-Free at time point $t$. NP-Free always uses three data points to predict the next data point. NP-Free trains its first LSTM model at $t = 2$ with data points $c_0$, $c_1$, and $c_2$ as input. It then uses this model to predict the next data point, $\widehat{c_3}$. The process repeats as $t$ advances, training new LSTM models and making predictions based on the three most recent data points.

At $t = 5$, NP-Free computes the prediction error using the well-known prediction accuracy metric, Root-Mean-Square Error (RMSE for short), as shown in Equation 2.

$$RMSE_t = \sqrt{\frac{\sum_{z=t-2}^{t}(c_z - \widehat{c_z})^2}{3}}, t \geq 5 \quad (2)$$

It is clear that given any two time series (says $A$ and $B$), RMSE is to evaluate the absolute error between $A$ and $B$, meaning that the error would be always identical no matter $A$ is the observed time series or the predicted one. That is why NP-Free has the characteristic of invariance.

After deriving $RMSE_5$, NP-Free continues to predict $\widehat{c_6}$ (see lines 9 and 10 of Figure 1). When $t$ equals 6, NP-Free repeats the same procedure to calculate $RMSE_6$ and predict $\widehat{c_7}$. When $t$ equals 7, NP-Free calculates $RMSE_7$, and it can also calculate $thd_{RMSE}$ by Equation 3.

```
NP-Free algorithm
Input: Each data point of the target time series
Output: A RMSE value
Procedure:
1:   Let t be the current time point and t starts from 0; Let Flag be True;
2:   While time has advanced {
3:       Collect data point c_t;
4:       if t ≥ 2 and t < 5 {
5:           Train an LSTM model by taking c_{t-2}, c_{t-1}, and c_t as the training data;
6:           Let m be the resulting LSTM model and use m to predict ĉ_{t+1};}
7:       else if t ≥ 5 and t < 7 {
8:           Calculate RMSE_t based on Equation 2 and output RMSE_t;
9:           Train an LSTM model by taking c_{t-2}, c_{t-1}, and c_t as the training data;
10:          Let m be the resulting LSTM model and use m to predict ĉ_{t+1};}
11:      else if t ≥ 7 and Flag==True {
12:          if t ≠ 7 { Use m to predict ĉ_t;}
13:          Calculate RMSE_t based on Equation 2;
14:          Calculate thd_{RMSE} based on Equation 3;
15:          if RMSE_t ≤ thd_{RMSE} { Output RMSE_t;}
16:          else{
17:              Train an LSTM model with c_{t-3}, c_{t-2}, and c_{t-1};
18:              Use the newly trained LSTM model to predict ĉ_t;
19:              Calculate RMSE_t based on Equation 2;
20:              Calculate thd_{RMSE} based on Equation 3;
21:              if RMSE_t ≤ thd_{RMSE} { Output RMSE_t;}
22:              else { Output RMSE_t; Let Flag be False;}}}
23:      else if t ≥ 7 and Flag==False {
24:          Train an LSTM model with c_{t-3}, c_{t-2}, and c_{t-1};
25:          Use the newly trained LSTM model to predict ĉ_t;
26:          Calculate RMSE_t based on Equation 2;
27:          Calculate thd_{RMSE} based on Equation 3;
28:          if RMSE_t ≤ thd_{RMSE} {
29:              Output RMSE_t;
30:              Replace m with the new LSTM model from line 24;
31:              Let Flag be True;}
32:          else { Output RMSE_t; Let Flag be False;}}}
```

Figure 1: The algorithm of NP-Free (Lee et al., 2023a).

$$thd_{RMSE} = \mu_{RMSE} + 3 \cdot \sigma \quad (3)$$

In Equation 3, $\mu_{RMSE}$ and $\sigma$ are the average RMSE value and the standard deviation at time point $t$, respectively. These values are calculated using Equations 4 and 5, respectively.

$$\mu_{RMSE} = \begin{cases} \frac{1}{t-4} \cdot \sum_{z=5}^{t} RMSE_z, & 7 \leq t < w+4 \\ \frac{1}{w} \cdot \sum_{z=t-w+1}^{t} RMSE_z, & t \geq w+4 \end{cases} \quad (4)$$

$$\sigma = \begin{cases} \sqrt{\frac{\sum_{z=5}^{t}(RMSE_z - \mu_{RMSE})^2}{t-4}}, & 7 \leq t < w+4 \\ \sqrt{\frac{\sum_{z=t-w+1}^{t}(RMSE_z - \mu_{RMSE})^2}{w}}, & t \geq w+4 \end{cases} \quad (5)$$

Here, $w$ is a parameter that limits the number of historical RMSE values considered when calculating the threshold. The purpose is to avoid exhausting system resources.

Whenever time point $t$ advances, and it is greater than or equal to 7 (i.e., either line 11 or line 23 of Figure 1 evaluates to true), NP-Free re-calculates $RMSE_t$ and $thd_{RMSE}$. If $RMSE_t$ is not greater than the threshold (as indicated in lines 15 and 28), NP-Free immediately outputs $RMSE_t$.

Otherwise, NP-Free tries to adapt to a potential pattern change by retraining an new LSTM model to re-predict $\widehat{c_t}$ and re-calculate both $RMSE_t$ and $thd_{RMSE}$ either at current time point (lines 17 to 20) or at the next time point (lines 24 to 27). If the re-calculated $RMSE_t$ is no larger than $thd_{RMSE}$, NP-Free immediately outputs $RMSE_t$. Otherwise, NP-Free outputs

$RMSE_t$ and performs LSTM model retraining at the next time point. This iterative process dynamically converts a time series into an RMSE series on the fly.

As stated earlier, NP-Free distinguishes itself from conventional representation approaches by eliminating preprocessing steps, including z-normalization. This characteristic enables NP-Free to serve as an alternative nomalization approach to z-normalization in clustering applications.

## 3 RELATED WORK

Kapil and Chawla (Kapil and Chawla, 2016) investigated the impact of different distance functions, including the Euclidean and Manhattan distance functions, on the performance of k-means. Ahmed et al. (Ahmed et al., 2020) conducted a review to address the shortcomings of the k-means algorithm, specifically focusing on the issues of initialization and its inability to handle data with mixed types of features. The authors performed an experimental analysis to investigate different versions of the k-means algorithms for different datasets.

Kuncheva and Vetrov (Kuncheva and Vetrov, 2006) investigated the stability of clustering algorithms, particularly cluster ensembles relying on k-means clusters, in the presence of random elements such as the target number of clusters (k) and random initialization. Using a diverse set of 10 artificial and 10 real datasets with a modest number of clusters and data points, the research assessed pairwise and non-pairwise stability metrics. The study explored the relationship between stability and accuracy concerning the number of clusters (k) and proposed a new combined stability index, incorporating both pairwise individual and ensemble stabilities, which shows improved correlation with ensemble accuracy.

Gupta and Chandra (Gupta and Chandra, 2020) aims to find out the possibility of different distance/similarity metrics to be used with k-means algorithm by conducting an empirical evaluation. The study compares the accuracy, performance, and reliability of 13 diverse distance or similarity measures across six variations of data using the k-means algorithm.

Vats and Sagar (Vats and Sagar, 2019) investigated the performance of the k-means algorithm through various implementations, including k-mean simple (utilizing Java codes on MapReduce), k-means with Initial Equidistant Centres (IEC), k-mean on Mahout (leveraging a machine learning library), and k-mean on Spark (utilizing another machine learning library). Additionally, the study explores the behavior of k-means algorithms concerning centroids and various iteration levels, providing insights into their performance across different infrastructures.

The work by Ikotun et al. (Ikotun et al., 2023) presents a comprehensive review focused on four key aspects: a systematic examination of the k-means clustering algorithm and its variants, the introduction of a novel taxonomy, in-depth analyses to validate findings, and identification of open issues. The review provides a detailed examination of k-means, identifies research gaps, and outlines future directions to address challenges in k-means clustering and its variants.

According to our investigation, there exists a gap in the literature concerning the evaluation of k-means time series clustering with different normalization techniques. This gap underscores the need for comprehensive studies that compare the performance of k-means clustering when utilizing various normalization techniques on time series data. Such investigations could provide valuable insights into the strengths and limitations of different normalization techniques and contribute to the development of more effective and robust clustering algorithms for time series data.

## 4 EVALUATION

To evaluate the impact of z-normalization and NP-Free on k-means time series clustering in terms of its clustering quality, we conducted two experiments using two real-world open-source time series datasets from the UEA&UCR archive (Dau et al., 2019). In the rest of this paper, we refer to k-means time series clustering based on z-normalization as **z-kmeans**, and refer to k-means time series clustering based on NP-Free as **NPF-kmeans**.

We implemented the above two variants using the k-means time series algorithm provided by tslearn (Tavenard et al., 2020), which is a Python package that provides machine learning tools for time series analysis. Furthermore, the NP-Free normalization in NPF-kmeans was implemented in DeepLearning4J (Deeplearning4j, 2023). To ensure a fair comparison, identical initial centroids were utilized for both z-kmeans and NPF-kmeans. All the experiments were performed on a laptop running MacOS Ventura 13.4 with 2.6 GHz 6-Core Intel Core i7 and 16GB DDR4 SDRAM.

In the first experiment, we selected all time series belonging to a class named "*class 2: Point (FP03, MP03, FP18, and MP18)*" from the GunPointAgeSpan_TRAIN.txt of the GunPointAgeS-pan dataset[1]. There are 67 raw time series in this class, with each time series representing an action performed by a person. We refer to this dataset as GunPointPointTrain in this paper. As per the dataset description, the time series can be categorized into various types. However, it is important to note that there is no label information indicating the specific type to which each time series belongs.

In the second experiment, we selected all time series from a class named "*class 2: Male (MG03, MP03, MG18, MP18)*" from the GunPointMaleVersusFemale_TRAIN.txt of the GunPointMaleVersusFemale dataset[2]. There are 64 raw time series in this class, and we denote this dataset as GunPointMaleTrain. Similarly, despite the dataset description suggesting that these time series can be further divided into different types, there is no label information in the dateset. However, these dataset are in their raw form, suitable for both z-normalization and NP-Free techniques. Table 1 summaries the characteristics of the two time series datasets.

Table 1: Summary of the two used open-source time series datasets.

| Name | Total number of time series | Length |
| --- | --- | --- |
| GunPointPointTrain | 67 | 150 |
| GunPointMaleTrain | 64 | 150 |

Table 2 presents the hyperparameter and parameter settings used by NP-Free in NPF-kmeans. This settings were originally suggested and employed in prior studies by (Lee et al., 2023a), (Lee et al., 2020), and (Lee et al., 2021). We have adopted these settings for our two experiments. Regarding the sliding window parameter $w$, it is suggested to be a large value (Lee and Lin, 2023). In this paper, $w$ was set to 150 because each time series in GunPointPointTrain and GunPointMaleTrain comprises only 150 data points.

Table 2: The hyperparameter and parameter settings used by NP-Free in this paper (Lee et al., 2023a).

| Hyperparameters and parameters | Value |
| --- | --- |
| The number of hidden layers | 1 |
| The number of hidden units | 10 |
| The number of epochs | 50 |
| Learning rate | 0.005 |
| Activation function | tanh |
| Random seed | 140 |

In order to evaluate the performance of z-kmeans and NPF-kmeans, Silhouettes (Rousseeuw, 1987) was used in this paper. Silhouettes is a well-known measure to evaluate the quality of clusters in unsupervised machine learning. Silhouettes quantifies how similar each object is to its own cluster compared to other clusters. A Silhouettes score ranges from -1 to 1. A value near 1 indicates that each object is well matched to its own cluster and poorly matched to neighboring clusters, a value of 0 indicates that each object is on or very close to the boundary between two neighboring clusters, and a value near -1 suggests that objects may have been assigned to the wrong cluster. The process for calculating the Silhouettes score to represent the overall clustering quality of a time series clustering approach is shown as follows:

1. Select a time series $i$: Choose one time series from a cluster for which we would like to calculate its Silhouettes score.

2. Calculate $a(i)$: Calculate the average distance of $i$ to all the other time series within the same cluster. A smaller value indicates a better assignment within the cluster.

3. Identify the nearest cluster to $i$: Compute the average distance between $i$ and all time series in each of the other clusters. Find the cluster that has the minimum average distance to $i$. This cluster is considered the nearest neighboring cluster to $i$.

4. Calculate $b(i)$: Calculate the average distance of $i$ to all time series in the nearest neighboring cluster.

5. Calculate $s(i)$: Calculate the Silhouettes score of $i$ using the following formula: $(b(i) - a(i))/max\{a(i), b(i)\}$. This resulting $s(i)$ ranges from -1 to 1, with higher values indicating better clustering.

6. Repeat steps 1 to 5 to calculate a Silhouettes score for each time series in the dataset. Afterward, calculate the average Silhouettes score for all the time series. This provides an overall measure of the clustering quality.

In this paper, we used the Silhouettes function provided by tslearn (Tavenard et al., 2020) to evaluate z-kmeans and NPF-kmeans. Note that the random state in tslearn is fixed at a value of 1 to avoid random execution results. Additionally, the metric for calculating the distance between time series is set to *euclidean*. It is worth mentioning that when evaluating NPF-kmeans, we mapped each RMSE series back to its original raw time series and calculated the average Silhouettes score using these raw time series. Similarly, we used the raw time series to compute the average Silhouettes score of z-kmeans. This is because using z-normalized time series to calcu-late the average Silhouettes score for z-kmeans unfairly favors z-kmeans over NPF-kmeans. Recall that z-normalization can make distinct time series seem similar, leading to them being assigned to the same cluster. Consequently, the overall Silhouettes score for z-kmeans might falsely appear higher than that of NPF-kmeans, giving the impression that z-kmeans provides better clustering quality.

---

[1] The GunPointAgeSpan dataset, http://www.timeseriesclassification.com/description.php?Dataset=GunPointAgeSpan

[2] The GunPointMaleVersusFemale dataset, http://www.timeseriesclassification.com/description.php?Dataset=GunPointMaleVersusFemale

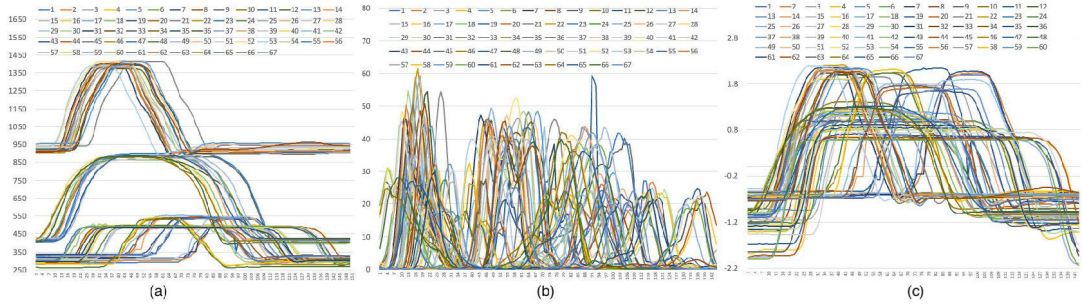

Figure 2: (a) The original raw time series in the GunPointPointTrain dataset, (b) The RMSE series of each raw time series, and (c) The z-normalized series of each raw time series.

Due to the mentioned reason, we evaluated the overall Silhouettes score of NPF-kmeans and z-kmeans using raw time series, which resulted in somewhat lower overall Silhouettes score. However, this approach ensures a fair basis for comparing NPF-kmeans and z-kmeans.

## 4.1 Experiment 1

In this experiment, we evaluated the individual performance of NPF-kmeans and z-kmeans on the GunPointPointTrain dataset. Figure 2(a) illustrates all the raw time series in this dataset. When NPF-kmeans was evaluated, it first converted each raw time series into a RMSE series using NP-Free. Figure 2(b) shows all the corresponding RMSE time series. On the other hand, when z-kmeans was evaluated, it first preprocessed each raw time series by translating it into a z-normalized series. The results can be seen in Figure 2(c). We then randomly chose 12 different values between 13 and 33 for the parameter $k$ in both NPF-kmeans and z-kmeans, with the aim of evaluating the overall Silhouettes score of the two approaches across different settings for $k$. As the results listed in Table 3, NPF-kmeans achieves a higher Silhouettes score than z-kmeans in all the cases. Note that all the scores were calculated based on raw time series, rather than normalized ones.

We further elaborate the clustering results of NPF-kmeans and z-kmeans when the GunPointPointTrain dataset was partitioned into 15 clusters, as both variants achieve the highest Silhouettes score under this setting. Figure 3 illustrates all the 15 clusters gener-ated by NPF-kmeans where the left part of the figure shows all the RMSE series in each cluster, whereas the right part of the figure shows all the corresponding raw time series in each cluster. It is clear that NPF-kmeans effectively clustered all the raw time series, as each time series within the same cluster is similar to every other but less similar to any time series in other clusters. It is important to note that this good performance cannot be reflected in the overall Silhouettes score of NPF-kmeans because the score was calculated based on raw time series rather than RMSE series. These scores were intended for the purpose of comparing NPF-kmeans and z-kmeans.

Table 3: The overall Silhouettes scores of NPF-kmeans and z-kmeans on GunPointPointTrain.

| The value of $k$ | NPF-kmeans | z-kmeans |
| --- | --- | --- |
| 13 | 0.4110 | 0.3327 |
| 14 | 0.4093 | 0.3654 |
| 15 | 0.5035 | 0.3766 |
| 17 | 0.4326 | 0.3594 |
| 19 | 0.3696 | 0.2853 |
| 20 | 0.3404 | 0.2845 |
| 21 | 0.3484 | 0.3129 |
| 22 | 0.3599 | 0.2787 |
| 26 | 0.3388 | 0.2345 |
| 28 | 0.3257 | 0.2004 |
| 29 | 0.3077 | 0.2332 |
| 33 | 0.2935 | 0.2294 |

Table 4: The clustering results of NPF-kmeans on GunPointPointTrain with $k$ set to 15.

| Cluster ID | # of time series | Time series No. |
| --- | --- | --- |
| 1 | 11 | 17,22,30,31,32,39,45,52,53,56,57 |
| 2 | 8 | 6,12,13,14,27,37,49,65 |
| 3 | 7 | 4,15,24,35,40,54,61 |
| 4 | 6 | 34,36,47,50,59,62 |
| 5 | 5 | 16,19,33,44,60 |
| 6 | 5 | 10,18,28,64,66 |
| 7 | 5 | 5,8,26,42,55 |
| 8 | 4 | 1,2,7,29 |
| 9 | 4 | 20,38,41,63 |
| 10 | 3 | 3,43,67 |
| 11 | 3 | 46,48,58 |
| 12 | 2 | 9,23 |
| 13 | 2 | 25,60 |
| 14 | 1 | 11 |
| 15 | 1 | 21 |

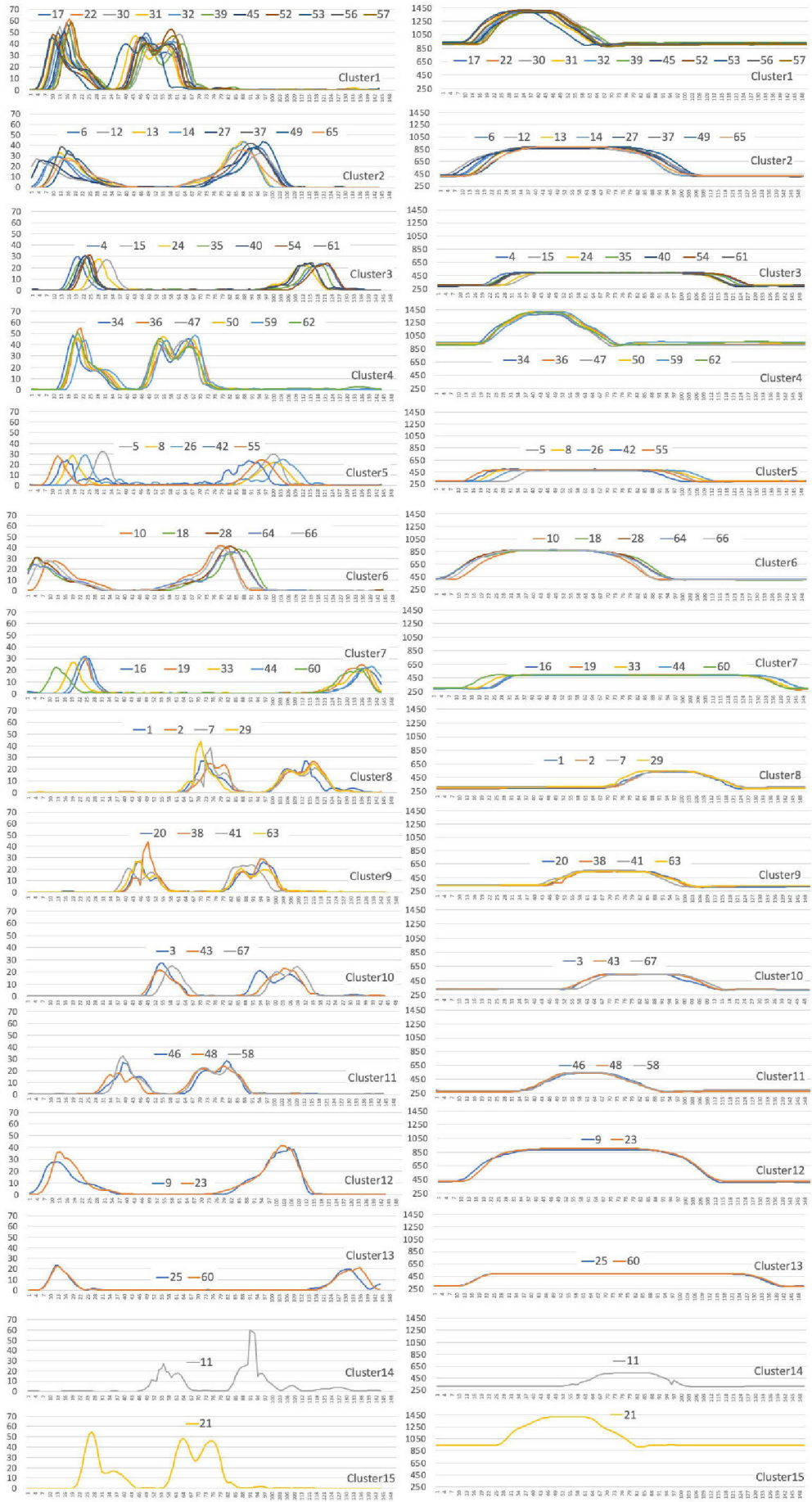

Figure 3: The clustering results of NPF-kmeans on the GunPointPointTrain dataset. The left part displays RMSE series in each cluster, while the right part shows the corresponding raw time series.

Table 4 further lists the detailed clustering results of NPF-kmeans. NPF-kmeans identified two time series as outliers (i.e., time series No. 11 and No. 21) and therefore assigned each of them to a separated cluster because their RMSE series are very different from those of the rest time series, which can be observed from the left part of Figure 3.

On the other hand, Figure 4 depicts all the 15 clusters produced by z-kmeans. The left part of the figure shows all the z-normalized time series in each cluster, while the right part shows all the corresponding raw time series. Detailed clustering results of z-kmeans are presented in Table 5. It is interesting and surprising to observe that, although the z-normalized time series in each cluster appear similar (as shown in the left part of Figure 4), not all corresponding raw time series within these clusters exhibit the same level of similarity (see the the right part of Figure 4). For instance, we can see that all z-normalized time series in cluster 2 appear significantly similar, but the corresponding raw time series in cluster 2 do not exhibit the same phenomenon. Apparently, raw time series No.42 and No.55 have a similar pattern with each other, but they have a much flatter pattern and much smaller y-axis values than the rest of the raw time series within cluster 2. Similar observations can be made in cluster 8.

The observed phenomena were attributed to the effects of z-normalization, which has the potential to make distinct time series indistinguishable. This finding is consistent with studies (Höppner, 2014) and (Lee et al., 2023a), suggesting that z-normalization might compromise potentially relevant properties that differentiate time series. Consequently, this could negatively impact subsequent data mining tasks (Dau et al., 2019; Senin, 2016; Codecademy-Team, 2022).

In this experiment, z-normalization misled time series to be wrongly clustered together into the same cluster. This is also why we chose not to use all the z-normalized time series for calculating the overall Silhouettes score of z-kmeans. The score, in this case, is unable to accurately reflect the true clustering quality of z-kmeans.

In terms of time consumption for NPF-kmeans and z-kmeans, we exclusively evaluated both variants for their preprocessing stages, as this is the only difference between them. Table 6 lists the average time consumption and standard deviation for NPF-kmeans and z-kmeans in preprocessing each raw time series of GunPointPointTrain using NP-Free and z-normalization, respectively. Since NP-Free is based on LSTM to generate a RMSE series for each time series of 150 data points, it took a longer preprocessing time than z-normalization.

Table 5: The clustering results of z-kmeans on GunPointPointTrain wiht $k$ set to 15.

| Cluster ID | # of time series | Time series No. |
|---|---|---|
| 1 | 11 | 17,22,30,31,32,39,45,52,53,56,57 |
| 2 | 9 | 6,12,13,14,27,37,42,55,65 |
| 3 | 7 | 21,34,36,47,50,59,62 |
| 4 | 5 | 10,18,28,64,66 |
| 5 | 5 | 16,19,33,44,51 |
| 6 | 4 | 1,2,7,29 |
| 7 | 4 | 3,11,43,67 |
| 8 | 4 | 8,9,23,49 |
| 9 | 4 | 20,38,41,63 |
| 10 | 3 | 4,35,54 |
| 11 | 3 | 26,40,61 |
| 12 | 3 | 46,48,58 |
| 13 | 2 | 15, 24 |
| 14 | 2 | 25,60 |
| 15 | 1 | 5 |

Table 6: The preprocessing time of NPF-kmeans and z-kmeans for each raw time series of GunPointPointTrain.

| Approach | Average time (sec) | Standard deviation (sec) |
|---|---|---|
| NPF-kmeans | 5.575 | 1.545 |
| z-kmeans | 0.002 | 0.001 |

## 4.2 Experiment 2

In this experiment, we further evaluated the clustering performance of NPF-kmeans and z-kmeans on the GunPointMaleTrain dataset. Figure 5(a) illustrates all the raw time series in this dataset, Figure 5(b) shows the corresponding RMSE time series generated by NP-Free, and Figure 5(c) depicts the corresponding time series generated by z-normalization.

Here we randomly selected 9 different values for parameter $k$, ranging from 13 to 29, for both NPF-kmeans and z-kmeans. This was done with the same purpose as mentioned in the first experiment: to evaluate the overall Silhouettes scores of the two approaches across different settings of $k$. As the results shown in Table 7, NPF-kmeans provides a higher overall Silhouettes score than z-kmeans in all the cases. Please note that all the Silhouettes scores shown in Table 7 might appear low, as they were calculated using the raw time series. As mentioned earlier that the scores were meant to compare NPF-kmeans and z-kmeans, so they cannot be used to present both variants' true clustering performance. It is evident from Table 7 that NPF-kmeans offers bet-ter clustering quality than z-kmeans, regardless of the value of $k$.

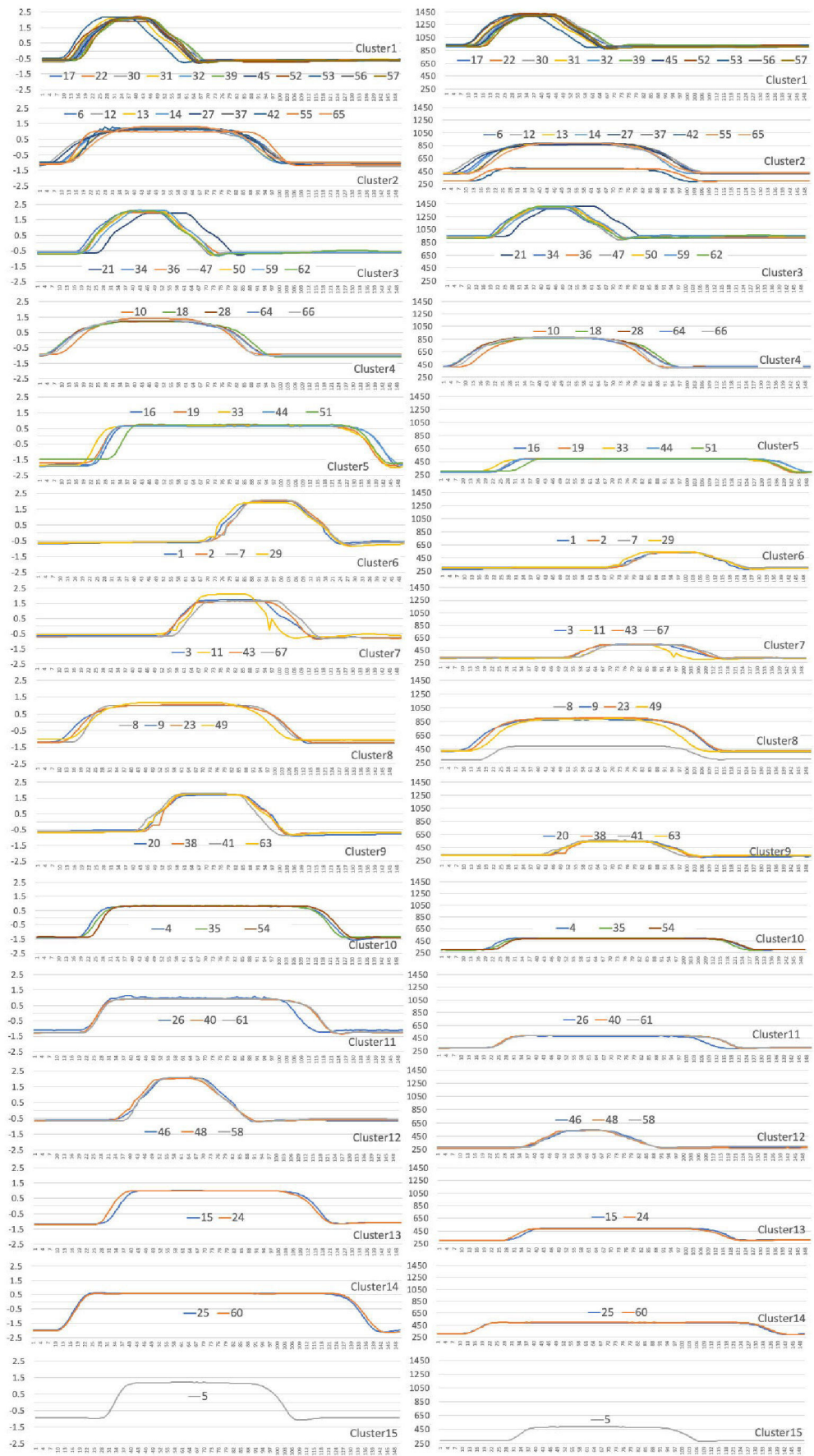

Figure 4: The clustering results of z-kmeans on the GunPointPointTrain dataset. The left part shows z-normalized time series in each cluster, while the right part shows the corresponding raw time series.

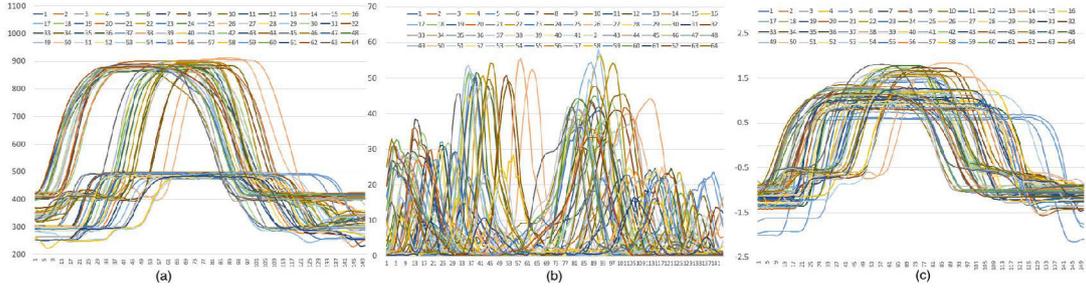

Figure 5: (a) The original raw time series in the GunPointMaleTrain dataset, (b) The RMSE series of each raw time series generated by NP-Free, and (c) The z-normalized series of each raw time series generated by z-normalization.

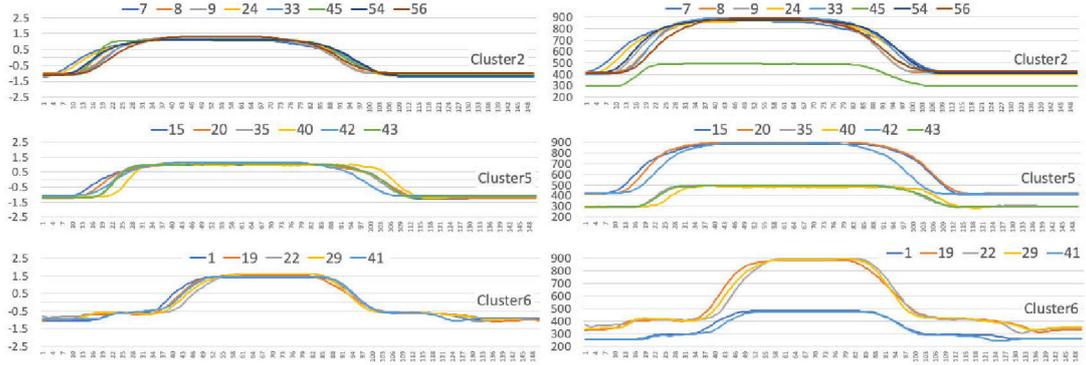

Figure 6: The clustering results of z-kmeans on the GunPointMaleTrain dataset. The left part shows all the z-normalized time series in clusters 2, 5, and 6, while the right part shows all the corresponding raw time series.

Table 7: The overall Silhouettes score of NPF-kmeans and z-kmeans on GunPointMaleTrain.

| The value of $k$ | NPF-kmeans | z-kmeans |
| --- | --- | --- |
| 13 | 0.2240 | 0.1512 |
| 15 | 0.2590 | 0.1986 |
| 16 | 0.3659 | 0.2231 |
| 18 | 0.3335 | 0.2284 |
| 20 | 0.3422 | 0.2108 |
| 23 | 0.2873 | 0.1858 |
| 26 | 0.2969 | 0.1339 |
| 28 | 0.2734 | 0.2085 |
| 29 | 0.2423 | 0.1991 |

To understand why z-kmeans performed worse than NPF-kmeans, we closely examined the clustering results of z-kmeans on the GunPointMaleTrain dataset, taking k =16 as an example. The left part of Figure 6 illustrates all the z-normalized time series in clusters 2, 5, and 6, while the right part of the same Figure depicts all the corresponding raw time series in these clusters. It is evident that all the z-normalized time series within each of these clusters are close to each other. This is why z-kmeans assigned these z-normalized time series to their respective clusters. However, if we map these z-normalized time series back to their original raw time series, we can see that not all the time series within each of these three clusters should be grouped together (see the right part of Figure 6). In other words, z-kmeans was misled by z-normalization.

Table 8 presents the average preprocessing time and standard deviation of NPF-kmeans and z-kmeans for each raw time series of GunPointMaleTrain. Similar to the result shown in the first experiment, NPF-kmeans required more preprocessing time than z-kmeans due to the adoption of NP-Free.

Table 8: The preprocessing time of NPF-kmeans and z-kmeans for each raw time series of GunPointMaleTrain.

| Approach | Average time (sec) | Standard deviation (sec) |
| --- | --- | --- |
| NPF-kmeans | 5.515 | 1.378 |
| z-kmeans | 0.002 | 0.001 |

## 5 CONCLUSIONS AND FUTURE WORK

In this study, we assessed the impact of utilizing z-normalization and NP-Free as normalization techniques on the performance of k-means time series clustering. Two experiments were conducted using two real-world open-source time series datasets.

Our findings indicate that NPF-kmeans (k-means based on NP-Free normalization) exhibited superior clustering results compared to z-kmeans (k-means based on z-normalization). The distinct advantage of NPF-kmeans lies in its ability to provide a more faithful representation of time series, which addresses concerns associated with the potential misguidance of z-normalization observed in z-kmeans.

However, our findings also indicate that the preprocessing part in NPF-kmeans requires a longer time due to the adoption of NP-Free, compared to z-kmeans. Therefore, to enhance the efficiency of NPF-kmeans when handling a large set of time series, it is recommended to integrate NPF-kmeans with parallelization in a multi-core environment or distributed computing clusters. This integration facilitates the acceleration of the preprocessing step, making NPF-kmeans more scalable. Additionally, it is also recommended to further enhance the performance of NP-Free, aiming to reduce the required time for converting a time series into an RMSE series.

In our future work, we plan to expand our evaluation by considering different time series clustering algorithms, different normalization techniques, and different performance metrics to provide a holistic and comprehensive evaluation.